\documentclass[sigconf]{acmart}
\usepackage{url}
\usepackage{booktabs}
\usepackage{tabularx}
\usepackage{caption}
\usepackage{multirow}
\usepackage{algorithm}
\usepackage{algorithmic}
\usepackage{subfigure}
\usepackage{amsmath}
\usepackage{enumitem}

\graphicspath{{01fig/}}

\AtBeginDocument{
    
}

\begin{document}

\setcopyright{acmlicensed}
\copyrightyear{2018}
\acmYear{2018}
\acmDOI{XXXXXXX.XXXXXXX}

\acmConference[Conference acronym 'XX]{Make sure to enter the correct
  conference title from your rights confirmation emai}{June 03--05,
  2018}{Woodstock, NY}
  
\acmISBN{978-1-4503-XXXX-X/18/06}

\title{Adaptive$^2$: Adaptive Domain Mining for Fine-grained Domain Adaptation Modeling}

\author{Wenxuan Sun*}
\affiliation{
 \institution{Peking University}
 \city{Beijing}
 \country{China}
}\email{sunwenxuan@stu.pku.edu.cn}

\author{Zixuan Yang*}
\affiliation{
 \institution{Beijing University of Posts and Telecommunications}
 \city{Beijing}
 \country{China}
}
\thanks{*Equal contribution. The work was done during their internship at Kuaishou.}
\email{shuaixuan666@bupt.edu.cn}

\author{Yunli Wang}
\affiliation{
 \institution{Kuaishou Technology}
 \city{Beijing}
 \country{China}
}
\email{wangyunli@kuaishou.com}

\author{Zhen Zhang}
\affiliation{
 \institution{Kuaishou Technology}
 \city{Beijing}
 \country{China}
}
\email{zhangzhen24@kuaishou.com}

\author{Zhiqiang Wang}
\affiliation{
 \institution{Kuaishou Technology}
 \city{Beijing}
 \country{China}
}
\email{wangzhiqiang03@kuaishou.com}

\author{Yu Li}
\affiliation{
 \institution{Kuaishou Technology}
 \city{Beijing}
 \country{China}
}
\email{liyu26@kuaishou.com}

\author{Jian Yang}
\affiliation{
 \institution{Beihang University}
 \city{Beijing}
 \country{China}
}
\email{jiaya@buaa.edu.cn}

\author{Yiming Yang}
\affiliation{
 \institution{Kuaishou Technology}
 \city{Beijing}
 \country{China}
}
\email{yangyiming03@kuaishou.com}

\author{Shiyang Wen}
\affiliation{
 \institution{Kuaishou Technology}
 \city{Beijing}
 \country{China}
}
\email{wenshiyang@kuaishou.com}

\author{Peng Jiang}
\affiliation{
 \institution{Kuaishou Technology}
 \city{Beijing}
 \country{China}
}
\email{jiangpeng@kuaishou.com}

\author{Kun Gai}
\affiliation{
 \institution{Independent}
 \city{Beijing}
 \country{China}
}
\email{gai.kun@qq.com}

\renewcommand{\shortauthors}{Wenxuan Sun et al.}

\begin{abstract}
  Advertising systems often face the multi-domain challenge, where data distributions vary significantly across scenarios. Existing domain adaptation methods primarily focus on building domain-adaptive neural networks but often rely on hand-crafted domain information, e.g., advertising placement, which may be sub-optimal. We think that fine-grained "domain" patterns exist that are difficult to hand-craft in online advertisement. Thus, we propose $Adaptive^2$, a novel framework that first learns domains adaptively using a domain mining module by self-supervision and then employs a shared\&specific network to model shared and conflicting information. As a practice, we use VQ-VAE as the domain mining module and conduct extensive experiments on public benchmarks. Results show that traditional domain adaptation methods with hand-crafted domains perform no better than single-domain models under fair FLOPS conditions, highlighting the importance of domain definition. In contrast, $Adaptive^2$ outperforms existing approaches, emphasizing the effectiveness of our method and the significance of domain mining. We also deployed $Adaptive^2$ in the live streaming scenario of Kuaishou Advertising System, demonstrating its commercial value and potential for automatic domain identification. To the best of our knowledge, $Adaptive^2$ is the first approach to automatically learn both domain identification and adaptation in online advertising, opening new research directions for this area.
\end{abstract}




\maketitle

\section{Introduction}

Online advertising systems have emerged as essential e-commerce marketing applications for targeting potential consumers of products or services, attracting substantial interest~\cite{cheng2016wide,mu2023hybrid,li2024deep,si2024twin,fan2024multi} and achieving significant advancements. 
To serve users' diverse interests and businesses' evolving needs, advertising systems typically incorporate data from multiple business scenarios, which exhibit both commonalities and diversities. The commonalities lie in the overlap between both users and items for different domains, while the diversities are caused by the different data distributions from multiple domains (e.g., ads' displaying placement, advertisers' marketing objective), leading to that user's preferences are quite various across domains \cite{sheng2021one}. 

Traditional advertising models have limitations in handling this multi-domain data \cite{yecsil2024star+}. They either consider all data samples independent and identically distributed from the same domain, or construct separate models for different scenarios \cite{jia2024d3}. The disadvantages of these approaches are obvious. On the one hand, learning from these combined datasets can make the model's task more difficult, as it fails to capture the commonalities and diversities, resulting in a notable loss of effectiveness. On the other hand, maintaining and training independent models for different scenarios can be resource-intensive and inefficient. 

Therefore, multi-domain models in advertising systems~\cite{ma2018entire,sheng2021one,jiang2022adaptive}, which can operate across different scenarios simultaneously, have been proposed to address these challenges. For instance, \citet{sheng2021one} have proposed the STAR model to tackle this problem, and \citet{jiang2022adaptive} have further extended it to the ADIN framework. These domain-adaptive methods employ a shared-and-specific architecture that integrates both domain-shared and domain-specific components within the model's main structure. Specifically, the domain-shared components are optimized using data from all domains, enabling the capture of common information shared across multiple scenarios. Conversely, the domain-specific components are designed to acquire information unique to each domain.

These methods have already demonstrated effective performance, but there still remain two challenges in the current models. On one hand, domain division \cite{jia2024d3} relies on predefined domain-related features, which are manually selected, the domain partitioning is fixed at the beginning of training. 
However, in real-world advertisement prediction tasks, numerous potential sub-domains have been identified, each characterized by a significant shift in data distribution \cite{zhang2024multi, zhang2024revisiting}. The manually selected domain features may be insufficient for the domain-specific network to capture these distinct data distribution differences. Moreover, the domain partitioning based on prior human knowledge may be sub-optimal for neural networks to learn the most relevant domain-specific representations. 
This highlights the need for more flexible and adaptive domain partitioning approaches. 

\begin{figure}[t]
    \centering
    \includegraphics[width=\linewidth]{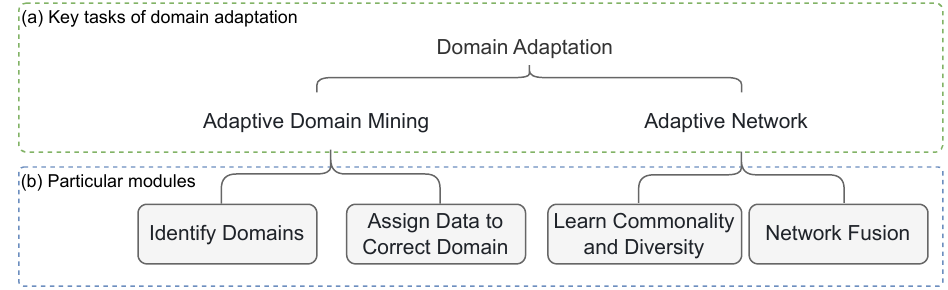}
    \caption{The key tasks of solving multi-domain problems and the corresponding components in Adaptive$^2$ are as above. Part (a) indicates the two aspects that domain adaptation should address, while part (b) shows the specific modules related to those aspects.}
    \label{key_task}
    \vspace{-5mm}
\end{figure}

Inspired by the exploration of previous work, we identify the key tasks in multi-domain advertising systems, as illustrated in Figure \ref{key_task} (a). The first challenge is to determine how many domains exist in data and which domain the data belong to, while the second challenge is to effectively learn both the commonalities and the distinctiveness within the different data distributions. Current adaptive methods have made progress in addressing the second task, with efforts to improve the adaptive network.  
However, they have seldom scrutinized the critical issue of domain identification \cite{jia2024d3}. 
Thus, we carve up the key tasks (shown in Figure \ref{key_task}(b)) and present a method named Adaptive$^2$ to extend the aforementioned adaptive methods. Specifically, our Adaptive$^2$ approach focuses on domain identification and leveraging domain information simultaneously by the Domain Mining Module and Adaptive Domain Modeling Module, respectively. 

The \textbf{Domain Mining Module} (DMM) addresses the challenge of domain identification. Previous methods often required manual annotations or expert knowledge to define the domains before training. However, empirically identifying all potential domains can be challenging, limiting the model's ability to leverage valuable domain information. We aim for an adaptive domain mining network that can uncover latent domains without the need for additional domain knowledge.
These subsets may exhibit certain conflicts in their data representations. 
Intuitively, self-supervised clustering is a suitable approach, thus we utilize VQ-VAE (Vector Quantized Variational Autoencoder) as a practical implementation. VQ-VAE classifies data into several learnable codebooks and reconstructs the original data through self-supervision. This process ensures that data points assigned to different codebooks come from distinct data distributions, while data points within the same codebook share more similar distributions. By treating each codebook as a domain, we can effectively identify and leverage latent domains in the data.

The \textbf{Adaptive Domain Modeling Module} (ADMM) focuses on commonality and diversity modeling and fusion. Our adaptive network inherits the shared and domain-specific structure of previous approaches, allowing it to adaptively learn both the common information and the domain-specific diversities.  
We route the samples into separate domain-specific networks, reducing the overall complexity and enabling the model to scale more effectively as the number of domains grows~\cite{fedus2022review}.

Extensive experiments are conducted on two public benchmarks of advertisement. The results show that $Adaptive^2$ outperforms all of the baselines. We also find that some baselines with hand-crafted domains perform no better than the single-domain model under relatively fair FLOPS settings, which indicates that domain definition for domain adaptation is quite important. We also conducted an ablation study to verify the impact of each proposed component, we can see that the VQ-VAE-based Domain Mining Module also has good generalization for other architectures (e.g., PEPNet), and the ADMM is a simple yet effective structure of domain adaptation. We also deploy $Adaptive^2$ to the Pre-ranking stage of the live streaming scenario in Kuaishou advertising system, bringing a 2.3\% increase in revenue and a 2.5\% increase in conversions.

Our main contributions can be summarized as follows: 1) We investigate the importance of domain identification for domain adaption in advertising system, and highlight that different domain adaptation methods should be compared from the perspective of resource fairness from an industrial perspective.
2) We propose the $Adaptive^2$ framework, which simultaneously achieves domain identification and adaptation, and we give a practice of $Adaptive^2$ based on VQ-VAE.
3) We conduct comprehensive offline and online experiments to verify the effectiveness of $Adaptive^2$. Experimental results show that $Adaptive^2$ surpasses the baselines with handcrafted domains, especially in the comparison under fairly FLOPs and parameter sizes, indicating the importance of domain identification. The result of online deployment in Kuaishou Technology shows a significant commercial value of $Adaptive^2$.

\begin{figure*}[t]
    \centering
    \includegraphics[width=0.77\linewidth]{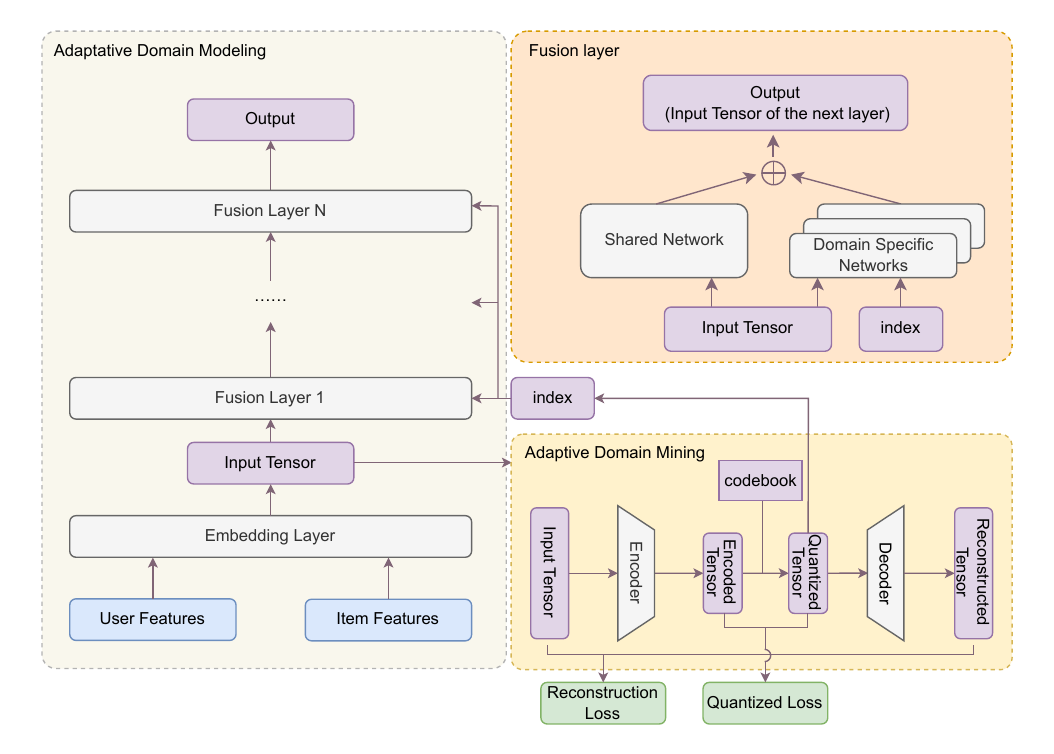}
    \caption{The Adaptive$^2$ framework, consists of the domain mining and the adaptive domain modeling module. The domain mining module outputs the domain index of each sample, which is used for routing to the domain-specific networks in the adaptive domain modeling module. Specifically, we employ VQVAE as the domain mining module.}
    \label{frame}
    \vspace{-3mm}
\end{figure*}

\section{Related Work}

\noindent\textbf{Domain Adaptation.} 
Domain Adaptation~\cite{ganin2015unsupervised, tzeng2017adversarial, zhao2022adpl, himeur2023video} is a branch of transfer learning, which leverages data from the source domain to enhance the performance of the target domain with limited labeled data. In domain adaptation problems, the data from the source and target domains have different distributions, but the tasks are the same. Because the independent and identically distributed (IID) assumption is not satisfied, models trained directly on the source domain usually perform poorly on the target domain. Thus, the key is how to transfer useful information from the source domain to the target domain. In the field of online advertising, a multitude of relevant endeavors have emerged in recent years, facilitating significant advancements in this area~\cite{sheng2021one,jiang2022adaptive,chang2023pepnet}. ADIN~\cite{jiang2022adaptive} uses domain-specific networks and shared networks to separately learn the commonalities and diversities of different domains. STAR~\cite{sheng2021one} proposes a star topology and a partitioned normalization method for transforming data distribution according to different domains. PEPNet~\cite{chang2023pepnet} proposes a Gate Neural Unit that maps the original embeddings to different domains, effectively addressing the semantic inconsistency in feature spaces across different domains. Despite the significant progress achieved by these works, specifying the number of domains requires additional expert knowledge. 

\noindent\textbf{Multi-task Learning.} Multi-task learning~\cite{ruder2017overview, kendall2018multi, ma2018modeling,bai2024multi} is a machine learning paradigm that simultaneously learns several related tasks by mining shared information, resulting in improved performance for each task. It has been widely used in various research fields, such as natural language processing~\cite{s2s_sls2020,gradVaccine2020,yangjian2022ganlm,yangjian2020alternating}, computer vision~\cite{uncertianty_weight2018,gradnorm2018}, online recommendation and advertising~\cite{tang2020progressive,ma2018modeling,metabalance2022}. Early widely used multi-task learning structures are shared-bottom architectures~\cite{caruana1997multitask} where the bottom hidden layers are shared to learn similar information, and the top layers use task-specific networks to learn different information. Instead of sharing hidden layers and identical model parameters across tasks, other methods use attention mechanisms and gating mechanisms to integrate information from different tasks. For instance, MTAN~\cite{liu2019end} uses attention modules to generate element masks, extracting task-specific knowledge from shared representations. Mixture-of-Experts (MoE)~\cite{jacobs1991adaptive} uses a gating mechanism to integrate information from task-specific experts to obtain the final result. As an extension of MoE, Multi-gate Mixture-of-Experts (MMoE)~\cite{ma2018modeling} employs multiple expert networks to capture various data patterns, enabling different tasks to utilize the experts in different ways. PLE~\cite{tang2020progressive} assigns independent experts for each task and considers the interactions between different experts to mitigate the seesaw phenomenon of MMoE. 

\vspace{2mm}
\noindent\textbf{Self-supervised Clustering.} Domain mining can be formulated as a self-supervised clustering task. Self-supervised learning is a powerful strategy that creates supervisory signals from the data itself, allowing models to learn useful representations without the need for extensive labeled datasets. This approach has proven particularly effective in extracting meaningful patterns that contribute significantly to downstream learning tasks\cite{stojnic2018analysis,mishra2020learning}, such as classification, object detection, and semantic segmentation.
VQ-VAE~\cite{van2017neural} learns intermediate codes through an encoder and then maps these intermediate codes to one of the K vectors in the codebook via nearest neighbor search which can be considered as clustering the intermediate codes. DeepCluster~\cite{caron2018deep} uses K-means clustering to generate pseudo-labels to guide convolutional network training. SwAV~\cite{caron2020unsupervised} introduces a clustering-based approach that performs online clustering while training, further pushing the boundaries of how clustering can be leveraged for self-supervised learning. \\
Some previous work has tried to alleviate these issues through improvements in modeling and sample construction~\cite{lee2022autoregressive}, but these approaches often come with high deployment costs and fail to comprehensively address both of the key challenges.

\section{Problem Formulation}

In this section, we first formulate the domain adaption ad ranking problem. Let $\mathcal{U}$ and $\mathcal{V}$ denote the collections of users and items respectively. $\mathbf{X, Y}$ signify the feature space and label space, with $\mathbf{X}_u$ representing user features and $\mathbf{X}_v$ representing item features, including both sparse and dense features like user behaviors.  The $n$-th domain is denoted as $\mathcal{D}_n$, and the data from domain $\mathcal{D}_n$ can be formulated as  $\{\mathbf{X}^n,\mathbf{Y}^n\}$, where $\mathbf{X}^n \subset \mathbf{X}$ represents the feature vectors of users and items in $\mathcal{D}_n$, and $\mathbf{Y}^n \subset \mathbf{Y}$ denotes the corresponding user behavior labels, such as click, conversion, etc. Formally, given the initial presentation $\mathbf{X}_u$ for the user and $\mathbf{X}_v$ for the item in the domain from different domains, we employ a model $\Phi$ parameterized by $\boldsymbol{\theta}$ to predict the class distribution as:
\begin{equation}
    \hat{\mathbf{Y}}^n = \Phi(\mathbf{X}^n_u,\mathbf{X}^n_v;\boldsymbol{\theta}),
\end{equation}
where $\hat{\mathbf{Y}}^n$ is the prediction in $\mathcal{D}_n$. Then, we use  Cross-Entropy (CE) as loss function and minimize the loss over all labeled node between  the ground truth and the prediction as
\begin{equation}
    \boldsymbol{\theta} = {\arg\min}_{\boldsymbol{\theta}}\mathcal{L} \text{, where }\mathcal{L} = \sum_{n=1}^N\textit{CE}(\mathbf{Y}^n,\hat{\mathbf{Y}}^n).
\end{equation}

\section{Approach}

In this section, we provide a detailed description of each component in our proposed method. The framework of our approach, denoted as Adaptive$^2$, is illustrated in Figure~\ref{frame}. 
Our method includes the following essential components: embedding layers for user and item features, a VQ-VAE-based adaptive domain mining module, a shared network, and domain-specific networks. 
First, user and item representations are obtained through the embedding layers and then combined into input tensors. These tensors are then processed by the VQ-VAE-based module for domain mining, which selects and utilizes the appropriate domain-specific networks based on the identified domains. 
Second, the outputs from the domain-specific networks and the shared network are then combined via a simple fusion layer to obtain the final representation.

\subsection{Self-supervised Adaptive Fine-grained Domain Mining}

In previous methods, domain information has to be annotated in advance, often requiring expert knowledge in the fields of recommendation or advertising. Identifying all domains empirically is often challenging, preventing the model from leveraging potential domain information to enhance performance. In this section, we present a self-supervised domain mining method that can uncover latent domains without the need for additional knowledge.\\
\noindent
\textbf{Embedding Layer.}
For the first step, we utilize an embedding layer to map diverse sparse features into a unified feature space, thereby deriving user and item features:
\begin{eqnarray}
\mathbf{x}_i = \textit{EMBED}(x_i) \\
\mathbf{x} = \textit{concat}(\mathbf{x}_1|...|\mathbf{x}_n),
\end{eqnarray}
where $x_i \in \mathbb{N}$  is the feature indicator of the $i_{\textrm{th}}$ feature slot, and $\mathbf{x}_i \in \mathbb{R}^d$ denotes $i_{\textrm{th}}$ embedding feature, with $d$ is the number of embedding dimensions. 
After we get the features of users $\mathbf{x}_u$ and items $\mathbf{x}_v$ through the embedding layer, we concatenate the features of pairs that have interacted, and subsequently process them through a Feed-Forward Network (FFN):
\begin{eqnarray}
    \mathbf{z} = \textit{FFN}(\textit{concat}(\mathbf{x}_u|\mathbf{x}_v))
\end{eqnarray}
\noindent
\textbf{Adaptive Domain Mining.}
In the following, we perform adaptive clustering to discover latent domain information. We utilize VQ-VAE~\cite{van2017neural}, a variant of the Variational Autoencoder (VAE)~\cite{Kingma2013AutoEncodingVB}, to mine latent domain information within the data. 
It employs a discrete latent space, known as a codebook, enhancing the generative model's expressiveness and stability in representing and generating data. Specifically, the input vector $\mathbf{z}$ is mapped to a latent representation through the encoder:
\begin{eqnarray}
\label{eq:encoder}
    \mathbf{z}_e = \textit{Encoder}(\mathbf{z}).
\end{eqnarray}
Then, during the vector quantization step, the continuous latent vectors produced by the encoder are mapped to the nearest discrete codebook vectors. Assume we have a codebook containing m embedding vectors, denoted as $\mathbf{E}$:
\begin{eqnarray}
    \mathbf{E} = \{\mathbf{e}_1,\mathbf{e}_2,...,\mathbf{e}_m\},
\end{eqnarray}
where $\mathbf{E} \in \mathbb{R}^{m\times d}$ and we consider $m$ is the number of latent domains.
For each $\mathbf{z}_e$, we compute its Euclidean distance to all codebook vectors and identify the nearest codebook vector. We obtain the corresponding codebook indicator $k$, which will be employed by the domain-specific networks as the domain indicator following the Vector Quantization step:
\begin{eqnarray}
    \mathbf{z}_q = \mathbf{e}_k, \quad \text{where}\quad k = {\arg\min}_j||\mathbf{z}_e-\mathbf{e}_j||_2.
\end{eqnarray}
Then, the quantized vector $\mathbf{z}_q$ is mapped back to the original data space via the decoder:
\begin{eqnarray}
    \hat{\mathbf{z}} = \textit{Decoder}(\mathbf{z}_q).
\end{eqnarray}

The loss function of VQ-VAE comprises three components: reconstruction loss, which measures the difference between the original input $\mathbf{z}$ and the reconstructed output $\hat{\mathbf{z}}$; vector quantization loss, which ensures the encoder output $\mathbf{z}_e$ is closer to the selected codebook vector $\mathbf{e}_{k}$; and batch regularization loss, which prevents the imbalanced usage of codebook vectors:
\begin{eqnarray}
    \mathcal{L}_{d} = ||\mathbf{z}-\hat{\mathbf{z}}||^2_2+||\rm sg(\mathbf{z}_e)-\mathbf{e}_{k}||_2^2+\beta||\mathbf{z}_e-\rm sg(\mathbf{e}_{k})||^2_2
\end{eqnarray}
where $\rm sg(\cdot)$ stands for the stop gradient operator, and $\beta$ is the commitment cost that helps balance the results of reconstruction and the flexibility of quantization.

It is important to note that our VQ-VAE-based domain mining method does not directly modify the input feature vectors. In practice, we apply the VQ-VAE technique on a vector with the gradient stopped, and use the resulting cluster assignments (domain indicator $k$) to guide the selection and utilization of domain-specific networks. This approach is adopted to avoid potential instability that the VQ-VAE method may introduce to the original input embeddings. By decoupling the domain mining and the feature encoding, we aim to ensure the robustness and reliability of the overall domain adaptation framework.

In summary, we achieve adaptive domain mining using VQ-VAE, with the discovered domain information represented in the form of a codebook. The domain information we discovered is embedded within the dataset and does not require additional domain expertise knowledge. The model architecture details of VQ-VAE's encoder and decoder are described in section~\ref{sec:exp_impl_details}.

\subsection{Fine-grained Domain Representation and Fusion}
In this section, we primarily present the fine-grained domain representation and fusion component of our model. 
Overall, our adaptive Fine-grained Domain Representation is divided into two components: Shared Network and Domain Specific Network. These components adaptively learn the common information in the data and the domain-specific diversities, respectively.\\
\noindent
\textbf{Shared Network.}
To learn the commonalities between different domains, we utilize a FFN (Feed-Forward Network)  shared in different domains to acquire the common representation:
\begin{eqnarray}
    \mathbf{O}_{\textrm{sh}} = \textit{FFN}(\mathbf{Z}),
\end{eqnarray}
where $\mathbf{Z}$ represents the input of FFN. The following uses the same notation.

\noindent
\textbf{Domain Specific Networks.}
We use Domain-Specific Networks to learn the diversities between different domains. 
Considering there are $N$ domains, we have $N$ such networks with different matrices:
\begin{eqnarray}
    \{\mathbf{W}_1,\mathbf{W}_2,...,\mathbf{W}_N\},
\end{eqnarray}
where each matrix represents the parameters of a specific domain network.
We use the domain indicator $k$ derived from the mined domain information to route the input tensors to different networks, obtaining the outputs from the corresponding domain-specific networks:
\begin{eqnarray}
    \mathbf{O}_{\textrm{sp}} = \mathbf{W}_{k}\mathbf{Z}.
\end{eqnarray}
In order to constrain the complexity, we only select the output of the most relevant domain network. However, we can also weight and fuse the outputs of all domain-specific networks based on their similarity to the codebook, as follows:
\begin{eqnarray}
    \begin{aligned}
        \alpha_j &= \textit{softmax }(f(\mathbf{z}_e,\mathbf{e}_j))\\
        \mathbf{O}_{\textrm{sp}} &= \sum_{j=1}^N \alpha_j \mathbf{W}_j \mathbf{Z},
    \end{aligned}
\end{eqnarray}
where $f(\cdot)$ denotes the similarity functions like cosine similarity and so on.

\noindent
\textbf{Fusion Layer.}
We simply add the shared representation $\mathbf{O}_{\textrm{sh}}$ and the domain-specific representation $\mathbf{O}_{\textrm{sp}}$ to obtain the fusion result. The resulting fused representation $\mathbf{O}_{\textrm{fusion}}$ can be formulated as Eq~\ref{eq:fusion}:

\begin{eqnarray}\label{eq:fusion}
    \mathbf{O}_{\textrm{fusion}} = \mathbf{O}_{\textrm{sh}} + \mathbf{O}_{\textrm{sp}}.
\end{eqnarray}

Note that the final output of the $Adaptive^2$ is the output of the final fusion layer, referring to "Fusion Layer N" in figure~\ref{frame}. After obtaining the final output of $Adaptive^2$, we can use a network $\Phi$ (reified as a linear regression in our experiments) parametrized by $\boldsymbol{\theta}$ to perform downstream tasks' (e.g, CTR prediction task as shown in Eq~\ref{eq:downstream_task}) predictions and compute the corresponding losses:
\begin{eqnarray}
    \hat{\mathbf{Y}} = \Phi(\mathbf{O}_{\textrm{fusion}};\boldsymbol{\theta})\\
    \mathcal{L}_{\textrm{task}} = \sum \textit{CE}(\mathbf{Y},\hat{\mathbf{Y}})\label{eq:downstream_task},
\end{eqnarray}
where $\mathbf{Y}$ denotes the ground truth. The total loss is the sum of the VQ-VAE loss and the downstream task loss:
\begin{eqnarray}
    \mathcal{L} = \mathcal{L}_{d} + \mathcal{L}_{\textrm{task}}.
\end{eqnarray}

\begin{algorithm}
\caption{Adaptive$^2$}
\label{alg:H2Gormer}
\begin{algorithmic}[1] 
\renewcommand{\algorithmicrequire}{\textbf{Input:}}
\renewcommand{\algorithmicensure}{\textbf{Output:}}

\REQUIRE User and item features $ x$ and labels $y$.
\ENSURE The Trained parameter $\boldsymbol{\theta}^*$.
\FOR{epoch from $1$ to $EPOCHS$}
    \STATE Obtain features $\mathbf{X}$ from the user and item embedding layer.
    \STATE Calculate feature projection $\mathbf{Z}$ using Eq.(5).
    \STATE Calculate common representations between different domains $\mathbf{O}_{\textrm{sh}}$ using Eq.(11).
    \STATE Apply stop gradient to the input of the VQ-VAE. Adaptive mining of the codebook id through VQ-VAE using Eq.(6) to Eq.(9).
    \STATE Obtain the domain specific network through ID routing and calculate the network output $\mathbf{O}_{\textrm{sp}}$ using Eq.(13) or Eq.(14).
    \STATE Combine $\mathbf{O}_{\textrm{sh}}$ and $\mathbf{O}_{\textrm{sp}}$ using Eq.(15)
    \STATE Calculate loss $\mathcal{L} = \mathcal{L}_{d} + \mathcal{L}_{\textrm{task}}$.
    \STATE Update the parameters $\boldsymbol{\theta}$ through back-propagation.
\ENDFOR
\end{algorithmic}
\end{algorithm}
\vspace{-3mm}
\subsection{Training Paradigm}
In advertising systems, streaming training is often employed for the sake of timeliness, meaning that the training data is drawn from a continuous data stream. There exist some trade-offs between co-training and pre-training VQ-VAE for the training paradigm. While a co-trained VQ-VAE may introduce some unstable factors for the training process compared to a pre-trained VQ-VAE, integrating it into the main model for joint training simplifies the training and deployment process. Considering that the instability introduced by co-training is more likely to slow down the initial convergence rather than affect the overall training process, and given that industrial advertising systems typically train for extended periods (e.g., several days to months) using streaming data, we opted for the co-training approach as shown in Algorithm~\ref{alg:H2Gormer} to keep the training and deployment process as simple as possible. During training, we apply stop gradient to the input embeddings of the VQ-VAE to ensure that the encoder, decoder, and codebook do not interfere with the embeddings learned by the shared\&specific networks.

\begin{table}[ht]
\centering
\caption{Statistics of datasets used in experiments.}
\resizebox{0.65\linewidth}{!}{
\begin{tabular}{cccc}
\toprule
Dataset & \#Interactions  & \#Features  & \#Fields \\
\midrule
Avazu & 40,428,967 & 1,544,250 & 22  \\
Criteo & 45,840,617 & 2,086,936 & 39 \\

\bottomrule
\end{tabular}
}
\label{dataset_statistic}
\end{table}

\begin{table}[t]
\caption{Experimental results on two datasets, where the bold indicates the best performance, and the underline indicates the second best. The italicized items in the table indicate statistically significant improvements over MLP (P-value<0.01).}
\centering
\resizebox{1.0\linewidth}{!}{
\begin{tabular}{llcccccc}
\toprule
\multirow{2}{*}{Datasets} & \multirow{2}{*}{Metrics} & \multicolumn{6}{c}{Models} \\
\cmidrule{3-8}
& & MLP &  MMoE & PLE & ADIN & PEPNet & Ours\\
\midrule
\multirow{2}{*}{Avazu} 
& AUC & 0.7759 & \emph{0.7816} & \emph{0.7821} & \emph{\underline{0.7823}} & \emph{0.7803}  & \textbf{\emph{0.7850}}\\
& LogLoss & 0.379  & 0.378 & 0.378 & 0.378 & 0.379 &  \textbf{\emph{0.376}}\\
\midrule
\multirow{2}{*}{Criteo} 
& AUC  & 0.8052  & \emph{0.8062} & \emph{\underline{0.8068}} & \emph{0.8067} & \emph{0.8061} &\textbf{\emph{0.8097}}\\
& LogLoss & 0.445  & 0.444 & 0.445 & 0.445 & 0.445 &  \textbf{\emph{0.441}}\\
\bottomrule
\end{tabular}
}
\label{tab:results}
\vspace{-1mm}
\end{table}

\section{Experiments}
In this section, we conduct extensive experiments to assess our model Adaptive$^2$ with other SOTA methods, aiming to answer the following four key questions. \\
\textbf{RQ1}: How does our model compare in performance to the current SOTA models in the advertisement recommendation?\\
\textbf{RQ2}: In our proposed method, how do the different components and implementations perform?  Can our mined domain information be directly applied to other methods to enhance their performance?\\
\textbf{RQ3}: How does our model perform compared to the current SOTA models under the Computational-Fair Comparison? \\
\textbf{RQ4}:How does our model perform in real online scenarios in Kuaishou advertising system?

\subsection{Offline Experiment Setup}
\subsubsection{Dataset Description}
To evaluate the effectiveness of our proposed method, we first conduct comprehensive experiments on two public CTR prediction benchmarks, namely Avazu\footnote{\url{https://www.kaggle.com/c/avazu-ctr-prediction}} and Criteo\footnote{\url{https://www.kaggle.com/c/criteo-display-ad-challenge}}, and then conduct online experiments on Kuaishou advertising system. The statistical data of the public benchmarks can be found in Table~\ref{dataset_statistic}. We utilize the preprocessed data from \cite{cheng2020adaptive} and follow the same settings for data splitting and preprocessing. The setups of online deployment are described in section~\ref{exp:online_deployment}.

\begin{figure}[ht]
    \centering
    \subfigure[Avazu]{\includegraphics[width=0.9\linewidth]{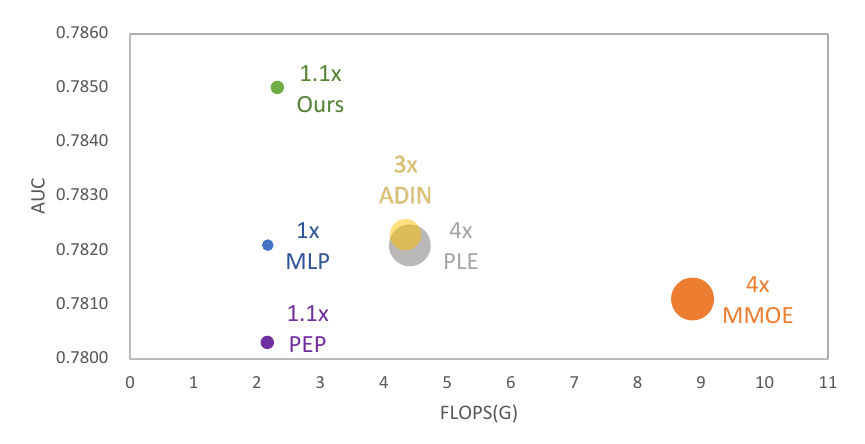}}
    \vspace{-4mm}
    \subfigure[Criteo]{\includegraphics[width=0.9\linewidth]{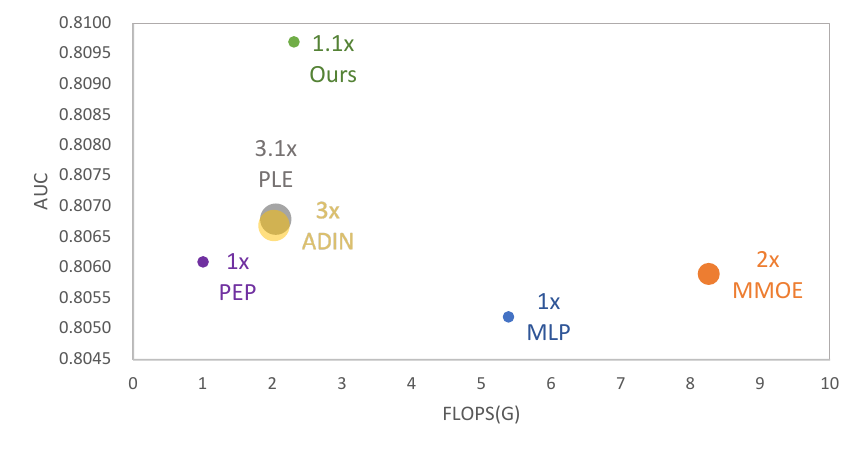}}
    \caption{Parameter comparison on two datasets. The horizontal axis represents computational FLOPs, the vertical axis represents AUC, and the numbers next to the models indicate the parameter count as a multiple of the MLP's parameters. The setting is identical to the main result.}
    \label{fig:parameters}
    \vspace{-3mm}
\end{figure}

\begin{table}[ht]
\caption{Ablation study results on two datasets, where ADMM is short for Adaptive Domain Modeling Module, DMM is short for Domain Mining Module, and HD($*$) means hand-crafted domains according to the feature $*$. $d1$ and $d2$ are carefully selected features suitable for domain info.}
\centering
\resizebox{\linewidth}{!}{
\begin{tabular}{llcccc}
\toprule
\multirow{2}{*}{Datasets} & \multirow{2}{*}{Metrics} & \multicolumn{4}{c}{Models} \\
\cmidrule{3-6}
&  & ADMM+HD(d1) & ADMM+HD(d2) & PEPNet+DMM & ADMM+DMM(Ours)\\
\midrule
\multirow{2}{*}{Avazu} 
& AUC & 0.7820 &  0.7816 & 0.7838  & 0.7850\\
& LogLoss & 0.378 & 0.378  & 0.376 &  0.376\\
\midrule
\multirow{2}{*}{Criteo} 
& AUC  & 0.8068 & 0.8036 & 0.8086 & 0.8097\\
& LogLoss & 0.445 & 0.449 & 0.443 & 0.441 \\
\bottomrule
\end{tabular}
}
\label{Ablation}
\vspace{-3mm}
\end{table}

\subsubsection{Evaluation Metric}
To evaluate the effectiveness of our method, we select various metrics to compare our approach with baselines. For the public datasets, we employ two widely-used evaluation metrics: AUC (Area Under ROC) and LogLoss (based on cross-entropy) which are important metrics in the CTR prediction task. Generally, a higher AUC or a lower Logloss value represents superior performance. For online experiments, we use the revenue and conversion numbers obtained after deploying the trained models to evaluate their performance.

\subsubsection{Competing Methods}
To demonstrate the effectiveness of our model, we select several state-of-the-art models for comparison. Detailed descriptions of these baseline models are as follows.

\begin{itemize}[leftmargin=*]
\item \textbf{MLP}~\cite{cheng2016wide} is the most commonly used deep learning model which comprises an embedding layer, multiple fully connected layers, and an output layer.

\item \textbf{MMoE}~\cite{ma2018modeling} utilizes multiple expert sub-models and a gating network shared across all tasks to implicitly capture the relationships among multiple tasks with distinct label spaces. 

\item \textbf{PLE}~\cite{tang2020progressive} establishes independent experts for each task and accounts for the interactions between experts while maintaining the shared experts from MMoE.

\item \textbf{ADIN}~\cite{jiang2022adaptive} introduces a domain interest adaptive layer that dynamically converts the original feature inputs into domain-specific features, and through shared and specialized networks, it models the commonalities and diversities of different domains.

\item \textbf{PEPNet}~\cite{chang2023pepnet} utilizes personalized prior information to improve embedding and parameter customization, and employs domain-specific towers exclusively for making predictions.
\end{itemize}

Note that ADIN and PEPNet are domain adaptation models designed specifically for online recommendation and advertising. These models take domain information as input. On the other hand, MMoE and PLE are multi-task models that use the same input for each domain. To incorporate domain information, we create variants of these models. Specifically, we treat each domain as a different task by using different experts in the output layer. In our experiments, we employ linear regression as the experts in the output layer. These experts are linear networks without normalization or activation functions, unlike the experts in the hidden layers.

\subsubsection{Implementation Details}\label{sec:exp_impl_details}
All models are implemented based on the Pytorch framework. We employ the AdamW optimizer~\cite{loshchilov2017decoupled} with an initial learning rate set at 1e-3 and a decay rate of 1e-4. We tune the learning rate from \{1e-3, 1e-4, 2e-4, 5e-4\} and dropout rate from \{0, 0.1, 0.2, 0.3, 0.4, 0.5\}. The embedding dimension is selected from \{32, 64\}, and the batch size for each dataset is 2048. We train all the models for 5 epochs with an early-stop strategy for the two public datasets. The number of Fusion Layers of $Adaptive^2$ is set to 3. The encoder and decoder of VQ-VAE are set to 3 layers of MLP. The number of codebooks is gradually increased until the performance stops improving. Each expert of the baseline models also adopts the MLP architecture. Each baseline model has 3 hidden expert layers. For all hidden layers of all models, we adopt batch normalization for their input and adopt PRelu~\cite{prelu2015} as the activation function. Besides, we tune the parameters of all baseline models on the validation set to ensure a fair comparison. For methods that require domain selection, we select the best result among the features.

\begin{table}[ht]
\caption{Comparison of different models on various metrics for Avazu and Criteo datasets under relatively fair FLOPS conditions between MLP and domain adaptation methods. The italicized items in the table indicate statistically significant improvements over MLP (P-value<0.01).}
\centering
\resizebox{\linewidth}{!}{
\begin{tabular}{llcccccc}
\toprule
\multirow{2}{*}{Dataset} & \multirow{2}{*}{Metrics} & \multicolumn{6}{c}{Model} \\
\cmidrule{3-8}
& & MLP  & MMoE & PLE & ADIN & PEPNet & Ours \\
\midrule
\multirow{2}{*}{Avazu} 
& AUC & 0.7759 & 0.7741 & 0.7749 & 0.7753 & 0.773  & \emph{0.7778} \\
& FLOPS & 1.43G & 1.346G & 1.17G & 1.096G & 1.43G  & 1.51G \\
\hline

\multirow{2}{*}{Criteo}
& AUC & 0.8048 & 0.804 & \emph{0.805} & 0.8049 & 0.8039  & \emph{0.8067} \\
& FLOPS &2.12G & 3.46G & 3.75G & 4.04G & 2.02G  & 2.17G \\
\bottomrule
\end{tabular}
}
\label{flops}
\end{table}

\subsection{Main Results (\textbf{RQ1})}\label{exp:main_results}

We compare the performance of our model Adaptive$^2$ and other baselines on two public datasets. The results are shown in Table~\ref{tab:results}. We have the following observations:
\begin{itemize}[leftmargin=*]
\item Our model Adaptive$^2$ surpasses all baseline models across all datasets and achieves competitive performances. Our model effectively mines the domain information needed for downstream tasks and learns both common and distinctive representations from these domain-specific networks. This strongly demonstrates the effectiveness of our model. 
\item The performance of multi-domain models (namely ADIN and PEPNet) generally surpasses the variants of multi-task models (namely MMoE and PLE), which should be attributed to the differences in their approaches to leveraging the domain info. This indicates that domain-routed experts would be better than the gated mechanism-routed experts in scenarios with explicit domains, where the gated mechanism is end-to-end learned.
\end{itemize}

We also record the FLOPs and parameter counts (evaluation details shown in appendix~\ref{app:flops}) for different models in Figure~\ref{fig:parameters}.  It is worth noting that the FLOPs of the MLP model differ significantly from those of other models. An unexpected finding is that a carefully tuned MLP model can achieve noteworthy performance. This observation highlights the importance of selecting the correct domain information. Unreasonable domain selection may introduce noise and produce counterproductive effects, further demonstrating the effectiveness of the domain information our method mined. More detailed analysis can be found in the section~\ref{exp:fairness}.

\subsection{Ablation Study (\textbf{RQ2})}\label{exp:ablation}
To investigate the role of the components of our method, we conduct an ablation study. In practice, we explore the effectiveness of adaptively clustering different domains using VQ-VAE. Firstly, we use fixed features as the domain identifiers instead of the identifiers obtained from the codebook of VQ-VAE clustering. Specifically, we select two features from the anonymous feature set, named d1 and d2. In terms of model architecture, we use our adaptive network called ADMM.
Secondly, we use PEPNet as the main network and utilize the output of VQ-VAE as the domain ids to select specific networks for different domains. This variant is named PEPNet+DMM (Domain Mining Module). The results can be found in Table~\ref{Ablation} and ADMM (Adaptive Domain Modeling Module)+DMM is the whole model of us. We obtain the following observations:
\begin{itemize}[leftmargin=*]
\item Methods with ours outperform others with fixed id across all datasets. This indicates that our proposed adaptive extraction method for domain information, which does not require additional domain knowledge or manual settings, performs better. It also demonstrates the effectiveness of our method.
\item Compared to other baseline models in Table~\ref{tab:results}, our method when integrated into ADMM and PEPNet achieves substantial improvements. This demonstrates that the domain information we mined achieves significant results in both simple networks such as ADMM and complex networks such as PEPNet. This further illustrates that adaptively mining domain information is superior to manually specified domains and does not require additional prior knowledge.
\item Our superior performance over PEPNet+DMM indicates that our model architecture not only differentiates the variations among different domains more effectively but also maintains the shared characteristics across domains.
\item The whole model Adaptive$^2$ can achieve the best performance, which demonstrates its effectiveness for adaptive domain mining for fine-grained domain adaptation modeling.
\end{itemize}
\subsection{Computational-Fair Comparison (RQ3)}\label{exp:fairness}
Previous studies rarely balance the complexity. The computational complexity of different baselines varies, as we set them according to previous work in domain adaptation, as shown in the Main Results (Section~\ref{exp:main_results}). However, in industrial applications, Return on Investment (ROI) is extremely important and there are several works in industrial scenarios to conduct model design and resource allocation around the ROI of machine resources~\cite{wang2024scaling,DCAF,CRAS,su2024rpaf}. Thus we further conduct a computational-fair comparison.
We measure computational and storage overhead using FLOPs and parameter counts, comparing models under nearly identical settings.
\\
\noindent\textbf{FLOPs-Fair Comparison.} It is well known that resources for computing FLOPs are often limited in industrial applications. If a new model performs better but requires a significant increase in machine resources, it is not cost-effective and not ideal in industrial application scenarios. Despite this, research focusing on equal computational resources, such as FLOPs, has been relatively scarce in the field of domain adaptation for online advertisements.
We evaluate the model’s performance under relatively close FLOPs, and the results are shown in Table~\ref{flops}. We have the following observations. The AUC metrics of the various models are relatively close to the Avazu and Criteo datasets. Among them, our model Adaptive$^2$ performs exceptionally well on the Avazu dataset, achieving an AUC of 0.7778, the highest among all models, demonstrating strong classification ability. Especially on the Criteo dataset, our Adaptive$^2$ model achieves the best performance under the smallest FLOPs setting, even surpassing an MLP model with twice the FLOPs. This is significant for resource efficiency, as it not only demonstrates that our model achieves state-of-the-art performance with the same FLOPs setting but also indicates that our model can achieve similar performance with fewer computational resources.

\noindent
\textbf{Parameter-Fair Comparison.} We quantify the number of parameters in different models, and the results are shown in Figure~\ref{fig:parameters}. The x-axis represents FLOPs, while the y-axis represents AUC. We use the parameter count of MLP as the baseline, with the numbers next to the models indicating how many times larger their parameters are compared to MLP. Additionally, larger icons indicate models with larger parameter sizes. From the results, we find that: 1) it can be observed that methods with performance similar to ours, such as ADIN and PLE, have parameter sizes approximately 2 to 3 times larger than ours. This indicates that our model is more lightweight while maintaining comparable effectiveness, and it also has greater potential for scaling; 2) with the same parameter size, our model significantly outperforms MLP and PEPNet in terms of both AUC and FLOPs. This demonstrates that our model is more effective in both the selection of domain information and the mining of representations across different domains. Overall, the relatively smaller parameter size implies reduced time consumption and also enhances the availability of Adaptive$^2$ in different settings.

To sum up, our model not only performs well under fair FLOPs constraints but also demonstrates excellent generalization ability under fair parameter size constraints.

\begin{table}[h]
\centering
\caption{Online experiment results. Each model is deployed with 10\% online traffic for 15 days.}
\resizebox{0.8\linewidth}{!}{
\begin{tabular}{|l|c|c|}
\hline
\textbf{Methods} & \textbf{Revenue} & \textbf{Conversion}  \\ \hline
Previous SOTA &   +0\% & +0\% \\ \hline
ADMM+HD(PageID) &  -0.112\% & -0.256\% \\
\hline
ADMM+DMM ($Adaptive^2$)  &  +2.376\%& +2.522\% \\
\hline
\end{tabular}
}
\vspace{-3mm}
\label{online}
\end{table}

\subsection{Online Deployment(\textbf{RQ4})}\label{exp:online_deployment}
To evaluate the impact of $Adaptive^2$ in real-world applications, we deployed the $Adaptive^2$ to the Pre-ranking stage of the live streaming scenario in Kuaishou advertising system.
The online base model adopts the multi-layer perceptrons (MLPs) architecture, which consists of 5 layers, whose output sizes are [1024,256,256,256,1]. Regarding features, we adopt both sparse and dense features to describe the information of users and ads in the online advertising system, including user profiles, user action lists, ad IDs, multi-modal ad embeddings, and so forth. We train the pre-ranking models using ARF~\cite{ARF} loss. The training objective is somewhat different from the CTR prediction of the public experiments, which further demonstrates the generalization capability of $Adaptive^2$ across diverse training objectives. We train all models under an online learning paradigm. We adopt the Adam optimizer and set the learning rate to 0.01. To fairly compare different methods, we cold-start train the models simultaneously. We put the models online for observation after a week of training to ensure they have converged. For online deployment, both the training and serving frameworks are developed based on TensorFlow\footnote{https://www.tensorflow.org}.

Note that the unified, domain-agnostic model represents our online state-of-the-art (SOTA) under fair machine resource comparisons. Methods like PLE and PEPNet, which we previously experimented with using ad placement as the domain, required significantly more machine resources to achieve better results. Thus, in the online experiment, we only compare $Adaptive^2$ to the online SOTA and the method (denoted as ADMM+HD(PageID)) that replaces the domain info of $Adaptive^2$ with ad placement. To ensure fairness in machine costs, we set the size of the ADMM encoder (MLP) to [128, 64] and the size of both the share and specific MLP networks to [448, 128, 128, 128, 1]. The decoder of VQ-VAE for online training is also an MLP model, with hidden units [64, 128]. We adopt PRelu~\cite{prelu2015} as the activation function and adopt batch normalization for hidden layers of all models.

The results of the online experiment are shown in Table~\ref{online}. We can see that $Adaptive^2$ has brought about a 2.3\% increase in advertising revenue and a 2.5\% increase in the number of user conversions. Such an improvement is considered significant in our advertising scenario. Additionally, the method with hand-crafted domains (refer to ad placement) does not surpass the online baseline, further indicating the importance of domain design.  

\section{Conclusion}
In this paper, we investigate the challenges of multi-domain scenarios and adaptive domain information mining.
We propose a novel approach called Adaptive$^2$, which comprises two key components: an adaptive domain mining module for directly extracting hidden domain information from the original data, and a shared \& specific network designed to learn both the similarities and differences across various domains. Extensive experiments on public benchmarks and online deployment in Kuaishou advertising system demonstrate the effectiveness of our proposed $Adaptive^2$. We further make a computational-fair comparison of different methods, highlighting the importance of domain mining and the effectiveness of $Adaptive^2$ in industrial scenarios.

\bibliographystyle{ACM-Reference-Format}
\bibliography{sample-base}

\appendix

\section{Evaluation Details of FLOPs and Parameter Counts}\label{app:flops}

Here we provide details on evaluating the FLOPs and parameter counts for the models using PyTorch. To ensure the accuracy of the evaluation, the parameter counts and FLOPs were calculated using the \textbf{profile} function of package \textbf{thop}. For FLOPs reported in the experiments, we measure the computational complexity based on training 4096 samples from the training set. Note that in our experiments, all samples have the same input embedding size, and the results based on 4096 samples can represent the overall performance on the dataset.

\section{Domain Visualization}

\begin{figure}[htp]
    \includegraphics[width=0.9\linewidth,height=6cm]{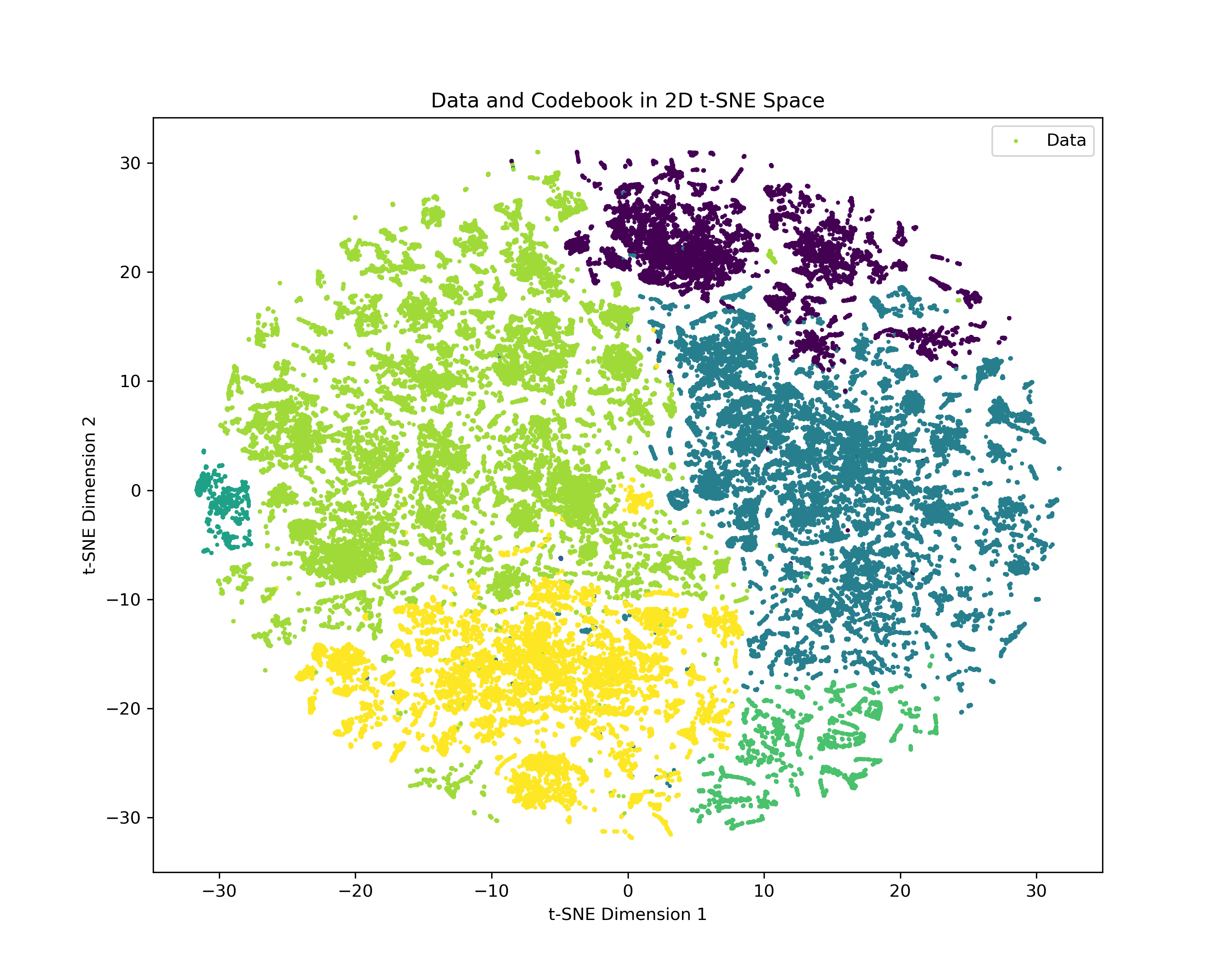}
    \vspace{-3mm}
    \caption{Visualization of the embeddings before VQ-VAE encoder on avazu.}
    \label{vis_emb_ori}
    \vspace{-3mm}
\end{figure}

\begin{figure}[htp]
    \includegraphics[width=0.9\linewidth,height=6cm]{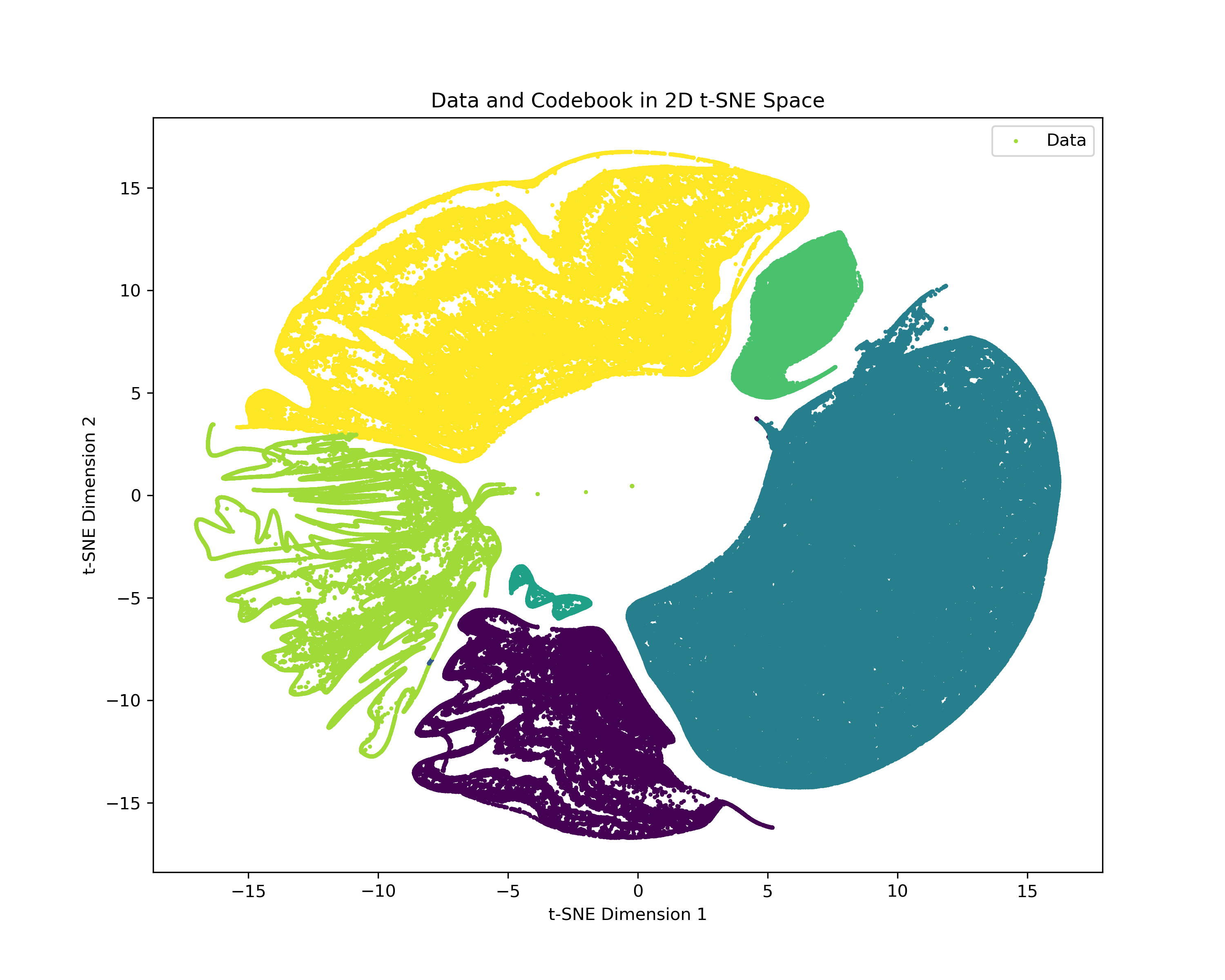}
    \vspace{-3mm}
    \caption{Visualization of the embeddings after VQ-VAE encoder on avazu.}
    \label{vis_emb}
    \vspace{-3mm}
\end{figure}

For a more intuitive comparison, we conduct the task of visualization on Avazu dataset.
We plot the output embedding before VQ-VAE's encoder ($\mathbf{z}$) and after it ($\mathbf{z}_e$) using t-SNE~\cite{lee2022autoregressive}, as defined in equation \eqref{eq:encoder}.
All nodes in Fig.~\ref{vis_emb_ori} and Fig.~\ref{vis_emb} are colored by the domain id. It can be observed that initially, the nodes are distributed throughout the entire space, with data from different domains interwoven. After mining, data within the same domain becomes closely clustered, and significant boundaries emerge between different domains. This further demonstrates that we have mined the implicit domain information present in the data.

\section{Analysis of VQ-VAE}

\begin{figure}[htp]
    \includegraphics[width=0.7\linewidth,height=5cm]{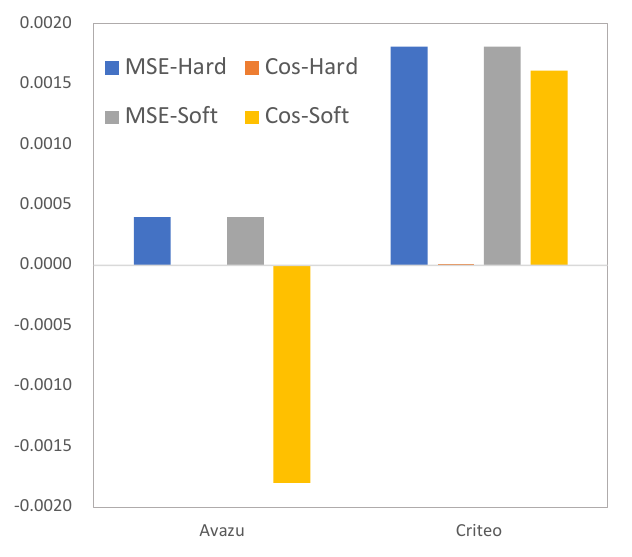}
    \caption{The analysis of different methods for obtaining domain id (MSE and Cos) and the domain-specific network selection (Hard and Soft). The vertical axis represents the difference between other methods and the Cos-Hard.}
    \label{A2-MSE}
    \vspace{-3mm}
\end{figure} 

We investigate the impact of VQ-VAE  various settings on adaptive domain mining.

Firstly, We explore two methods for obtaining the codebook ids in VQ-VAE: 1) calculating the mean-square error (MSE) between the encoder's output vector and the codebook vectors, referred to as MES and 2) using the cosine similarity between the encoder's output vector and the codebook vectors, called Cos. The results can be found in Fig.~\ref{A2-MSE}. Overall, among the two methods, the MES approach performs better. 
Secondly, we explore two methods for fusing domain-specific networks: 1) selecting the result of a single domain network based on the maximum similarity, referred to as Hard, and 2) weighted fusion of all domain networks based on softmax values, referred to as Soft. The results indicate that the performance of a single domain network surpasses that of the weighted fusion on both datasets. Meanwhile, this indicates that the domain information we mined is orthogonal, meaning that introducing weighted representations from other domains essentially introduces noise. This indirectly confirms that the domain information we mined is both efficient and meaningful.

In future work, we will also explore the impact of more detailed VQ-VAE optimizations on domain mining, such as preventing VQ-VAE collapse and investigating how different codebook definitions affect the mining results.

\section{Limitations and Future Work}
To the best of our knowledge, we are the first to propose an adaptive domain mining framework and implement a VQ-VAE-based mining method, achieving impressive results. However, our approach still has some limitations. Firstly, using VQ-VAE as the mining network is an empirical choice, as we believe it has stronger representational power. The impact of different self-supervised domain mining methods such as SwAV, and IDEC on domain mining and how to ensure the number of latent domains remains to be explored. 
Secondly, the domain information we mine outperforms previous SOTA methods in terms of effectiveness. However, compared to manually defined domain information with good interpretability, the implicit domains often lack direct real-world interpretability and sufficient theoretical support. Future work could explore how to enhance the supervision of domain mining from these two aspects. 

\end{document}